\documentclass[10pt,twocolumn,twoside]{IEEEtran}
\usepackage{amsmath,graphicx}
\usepackage{amssymb}
\usepackage{graphicx}
\usepackage{graphics}
\usepackage{epstopdf}
\usepackage{amsmath}
\usepackage{algorithmicx}
\usepackage{algorithm}
\usepackage{algpseudocode}
\usepackage{multirow}
\usepackage{bm}
\usepackage{enumerate}

\newcommand{\BigO}[1]{\ensuremath{\operatorname{O}\bigl(#1\bigr)}}

\begin{document}

\title{On learning with shift-invariant structures}

\author{Cristian Rusu\thanks{The author is with the Istituto Italiano di Tecnologia (IIT), Genova, Italy. Contact e-mail address: cristian.rusu@iit.it. Demo source code: https://github.com/cristian-rusu-research/shift-invariance}}

\maketitle

\begin{abstract}
In this paper, we describe new results and algorithms, based on circulant matrices, for the task of learning shift-invariant components from training data. We deal with the shift-invariant dictionary learning problem which we formulate using circulant and convolutional matrices (including unions of such matrices), define optimization problems that describe our goals and propose efficient ways to solve them. Based on these findings, we also show how to learn a wavelet-like dictionary from training data. We connect our work with various previous results from the literature and we show the effectiveness of our proposed algorithms using synthetic as well as real ECG signals and images.
\end{abstract}


\section{Introduction}

Circulant matrices \cite{Gray2005} are highly structured matrices where each column is a circular shift of the previous one. Because of their structure and their connection to the fast Fourier transform \cite{SR} (circulant matrices are diagonalized by the Fourier matrix \cite[Section 3]{Gray2005}), these matrices have seen many applications in the past: computing the shift between two signals (the shift retrieval problem exemplified in the circular convolution and cross-correlation theorems) for the GPS locking problem \cite{GPS}, time delay estimation \cite{TDE}, compressed shift retrieval from Fourier components \cite{CSR2}, matching or alignment problems for image processing \cite{Kanade}, designing numerically efficient linear transformations \cite{CircApprox} and overcomplete dictionaries \cite{ConvolutionalTree}, matrix decompositions \cite{Ye2016}, convolutional dictionary learning \cite{FCSC1,FCSC2,FCSC3} and sparse coding \cite{Elad2017}, learning shift invariant structured from data \cite{BeforeCDLA},\cite{CDLA} for medical imaging \cite{Medical2014}, EEG \cite{KSVDExtended} and audio \cite{AudioApplication} signal analysis. A recent review of the methods, solutions and applications related to circulant and convolutional representations is given in \cite{EfficientConvSparse}.

In this paper, we propose several numerically efficient algorithms to extract shift-invariant components or alignments from data using several structured dictionaries related to circulant matrices. 

Previously, several dictionary learning techniques that accommodate for shift invariance have been proposed: extending the well-known K-SVD algorithm to deal with shift-invariant structures \cite{KSVDExtended, AharonPhd2006, Spanias2008}, proposing a shift-invariant iterative least squares dictionary learning algorithm \cite{SIDLA}, extending the dictionary while solving an eigenvalue problem \cite{MOTIF}, fast online learning approach \cite{Zheng:2016:ESD:2939672.2939824}, research that combines shift and 2D rotation invariance \cite{Mars2012} and proposing new algorithms that optimize directly the dictionary learning objective functions with circulant matrices \cite{BeforeCDLA, CDLA, BARZIDEH2017162}. The convolutional sparse representation model \cite{NgShift-Invariant, Wetzstein2015, Lucey2014} where the dictionary is a concatenation of circulant matrices has been extensively studied in the past. Furthermore, recent work \cite{Elad2017} uses tools developed in the sparse representations literature to provide theoretical insights into convolutional sparse coding where the dictionary is a concatenation of banded circulant matrices and its connection to convolutional neural networks \cite{Neural2016}. Detailed literature reviews of these learning and convolutional sparse representations problems and proposed solutions have been recently described in 
\cite[Section II]{EfficientConvSparse}.

Structured dictionaries have received a lot of attention mainly because of two reasons: the structure means that the dictionaries will be easier to store and use (lower memory footprint and lower computational complexity to perform, for example, matrix-vector multiplication or solving linear systems) and they act as regularizers modeling some property of the data that is interesting. In the case of shift-invariant dictionaries these two advantages are transparent: manipulation of circulant matrices is done via the fast Fourier transform while storing them takes only linear space (instead of quadratic) and they are able to model patterns from the data that are repetitive, as we expect real-world data (especially image data and time-series) to exhibit such patterns.

We start by outlining in Section II circulant matrices and their properties, particularly their factorization with the Fourier matrix, and other structured matrices that we use in this paper. Then, we propose algorithms to learn shift-invariant (circulant, convolutional and unions of these) and wavelet-like components from training data (Section III). Finally, in Section IV, we show experimental results with various data sources (synthetic, ECG, images) that highlight the learning capabilities of the proposed methods.

\noindent\textbf{Notation}: bold lowercase $\mathbf{x} \in \mathbb{R}^n$ is used to denote a column vector, bold uppercase $\mathbf{X} \in \mathbb{R}^{n \times m}$ is used to denote a matrix, non-bold lowercase Greek letters like $\alpha \in \mathbb{R}$ are used to denote scalar values, calligraphic letters $\mathcal{K}$ are used to denote sets and $|\mathcal{K}|$ is the cardinality of $\mathcal{K}$ (abusing this notation, $|\alpha|$ is the magnitude of a scalar). Then $\| \mathbf{x} \|_2$ is the $\ell_2$ norm, $\| \mathbf{x} \|_{0}$ is the $\ell_0$ pseudo-norm,  $\| \mathbf{X} \|_F^2 = \text{tr}(\mathbf{X}^H \mathbf{X})$ is the Frobenius norm, $\mathrm{tr}(\mathbf{X})$ denotes the trace, $\text{vec}(\mathbf{X}) \in \mathbb{R}^{nm}$ vectorizes the matrix $\mathbf{X} \in \mathbb{R}^{n \times m}$ columnwise, $\text{diag}(\mathbf{x})$ denotes the diagonal matrix with the vector $\mathbf{x}$ on its diagonal, $\mathbf{X}^H$ is the complex conjugate transpose, $\mathbf{X}^T$ is the matrix transpose, $\mathbf{X}^*$ is the complex conjugate, $\mathbf{X}^{-1}$ denotes the inverse of a square matrix, $x_{kj}$ is the $(k,j)^{\text{th}}$ entry of $\mathbf{X}$. Tilde variables like $\mathbf{\tilde{X}}$ represents the columnwise Fourier transform of $\mathbf{X}$, $\mathbf{X} \otimes \mathbf{Y}$ denotes the Kronecker product and $\mathbf{X} \odot \mathbf{Y}$ is the Khatri-Rao product \cite{KhatriRao}.

\section{The proposed structured dictionaries}

\subsection{Circulant dictionaries}

We consider in this paper circulant matrices $\mathbf{C}$. These square matrices are completely defined by their first column $\mathbf{c} \in \mathbb{R}^n$: every column is a circular shift of the first one. With a down shift direction the right circulant matrices are:
\begin{equation}
\begin{aligned}
\mathbf{C} =  \text{circ}(\mathbf{c}) \stackrel{\rm def}{=} & \begin{bmatrix}
				c_1 & c_{n} & \dots & c_3 & c_2 \\
				c_2 & c_1 & \dots & c_4 & c_3 \\
				\vdots & \ddots & \ddots & \ddots & \vdots \\
				c_{n-1} & c_{n-2} & \dots & c_1 & c_n \\
				c_n & c_{n-1} & \dots & c_2 & c_1
				\end{bmatrix} \\
= & \begin{bmatrix} \mathbf{c} & \mathbf{P}\mathbf{c} & \mathbf{P}^2\mathbf{c} & \dots & \mathbf{P}^{n-1}\mathbf{c} \end{bmatrix} \in \mathbb{R}^{n \times n}.
\end{aligned}
	\label{eq:circulant}
\end{equation}

The matrix $\mathbf{P} \in \mathbb{R}^{n \times n}$ denotes the orthonormal circulant matrix that circularly shifts a target vector $\mathbf{c}$ by one position, i.e., $\mathbf{P} = \text{circ}(\mathbf{e}_2)$ where $\mathbf{e}_2$ is the second vector of the standard basis of $\mathbb{R}^{n}$. Notice that $\mathbf{P}^{q-1} = \text{circ}(\mathbf{e}_q)$ is also orthonormal circulant and denotes a cyclic shift by $q-1$. The main property of circulant matrices \eqref{eq:circulant} is their eigenvalue factorization which reads:
\begin{equation}
	\mathbf{C} = \mathbf{F}^H \mathbf{\Sigma} \mathbf{F},\ \mathbf{\Sigma} = \text{diag}(\mathbf{\sigma}) \in \mathbb{C}^{n \times n},
\label{eq:cfactorizationr}
\end{equation}
where $\mathbf{F} \in \mathbb{C}^{n \times n}$ is the unitary Fourier matrix ($\mathbf{F}^H\mathbf{F} = \mathbf{F}\mathbf{F}^H = \mathbf{I}$) and the diagonal $\mathbf{\sigma} = \sqrt{n}\mathbf{Fc},\ \mathbf{\sigma} \in \mathbb{C}^n$.

Note that the multiplication with $\mathbf{F}$ on the right is equivalent to the application of the Fast Fourier Transform, i.e., $\mathbf{Fc} = \text{FFT}(\mathbf{c})$, while the multiplication with $\mathbf{F}^H$ is equivalent to the inverse Fourier transform, i.e., $\mathbf{F}^H\mathbf{c} =  \text{IFFT}(\mathbf{c})$. Both transforms are applied in $\BigO{n \log n}$ time and memory. 

\subsection{Convolutional dictionaries}

Convolutional dictionaries can be reduced to circulant dictionaries by observing that given $\mathbf{c} \in \mathbb{R}^{n}$ and $\mathbf{x} \in \mathbb{R}^{m}$ with $m \geq n$ the result of their convolution is a vector $\mathbf{y}$ of size $p = n+m-1$ as
\begin{equation}
\begin{aligned}
\mathbf{y} = & \mathbf{c} * \mathbf{x} = \text{toep}(\mathbf{c}) \mathbf{x} = \begin{bmatrix}
c_1 & 0 & \dots & 0 & 0 \\
c_2 & c_1 & \dots & 0 & 0 \\
\vdots & \ddots & \ddots & \ddots & \vdots \\
c_n & c_{n-1} & \dots & c_1 & 0 \\
0 & c_n & \dots & c_{2} & c_1 \\
0 & 0 & \ddots & c_3 & c_{2} \\
\vdots & \vdots & \ddots & \ddots & \vdots \\
0 & 0 & \dots & 0 & c_{n}
\end{bmatrix} \mathbf{x} \\
= &
\text{circ} \left( \! \begin{bmatrix}
\mathbf{c} \\ \mathbf{0}_{(m-1) \times 1}
\end{bmatrix} \! \right) \begin{bmatrix} \mathbf{x} \\ \mathbf{0}_{(n-1) \times 1}
\end{bmatrix}  = \mathbf{C}_\text{conv} \mathbf{x}_\text{conv},
\end{aligned}
\label{eq:convolution}
\end{equation}
where $\text{toep}(\mathbf{c})$ is a Toeplitz matrix of size $p \times m$. Padding with zeros such that all variables are of size $p$ leads again to a matrix-vector multiplication by a circulant matrix. An alternative, but ultimately equivalent, way to write the convolution in terms of a circulant matrix is to notice that a Toeplitz matrix can be embedded into an extended circulant matrix of twice the size (see \cite[Section 4]{CDLA}).

For our purposes, there is a fundamental difference between $\mathbf{C}$ and $\mathbf{C}_\text{conv}$: in the case of $\mathbf{C}$ it is exactly equivalent if we choose to operate with $\mathbf{c}$ or $\mathbf{\sigma}$ while for $\mathbf{C}_\text{conv}$ we necessarily have to work with $\mathbf{c}_\text{conv}$ in order to impose its sparse structure. As we will see, this means that in the convolutional case we cannot exploit some simplifications that occur in the Fourier domain (when working directly with $\mathbf{\sigma}$).

\subsection{Wavelet-like dictionaries}

Multiples, powers, products, and sums of circulant matrices are themselves circulant. Therefore, extending this class of structures to include other dictionaries is not straightforward.

In order to represent a richer class of dictionaries, still based on circulants, consider now the following structured matrix
\begin{equation}
\begin{aligned}
\mathbf{C}_k^{(p)} 
= & \begin{bmatrix}
\mathbf{G}_k\mathbf{S} & \mathbf{H}_k \mathbf{S}
\end{bmatrix}
\in \mathbb{R}^{p \times p},
\end{aligned}
\label{eq:theWn}
\end{equation}
where $\mathbf{G}_k = \text{circ}(\mathbf{g}_k)$ and $\mathbf{H}_k = \text{circ}(\mathbf{h}_k)$ are both $p \times p$ circulant matrices and $\mathbf{S} \in \mathbb{R}^{p \times \frac{p}{2}}$ is a selection matrix that keeps only every other column, i.e., $\mathbf{G}_k \mathbf{S} = \begin{bmatrix}
\mathbf{g}_k & \mathbf{P}^2 \mathbf{g}_k & \mathbf{P}^4 \mathbf{g}_k & \dots & \mathbf{P}^{n-2} \mathbf{g}_k
\end{bmatrix} \in \mathbb{R}^{p \times \frac{p}{2}}$ (downsampling the columns by a factor of 2). In general we assume that the filters $\mathbf{g}_k$ and $\mathbf{h}_k$ have compact support with length denoted $n \leq p$. As such, these transformations are related to the convolutional dictionaries previously described. We call $\mathbf{g}_k$ and $\mathbf{h}_k$ filters because  \eqref{eq:convolution} is equivalent to a filtering operation of a signal $\mathbf{x}$ where the filter coefficients are stored in the circulant matrix.

Now we define a new transformation that operates only on the first $\frac{p}{2^{k-1}}$ coordinates and keeps the other unchanged:
\begin{equation}
\mathbf{W}_k^{(p)} \! \! = \! \! \begin{bmatrix}
\mathbf{C}_k^{\left( \frac{p}{2^{k-1}} \right)} & \! \mathbf{0}_{\frac{p}{2^{k-1}} \times \left( p - \frac{p}{2^{k-1}} \right) } \\
\mathbf{0}_{\left( p - \frac{p}{2^{k-1}} \right) \times \frac{p}{2^{k-1}}} & \mathbf{I}_{p - \frac{p}{2^{k-1}}}
\end{bmatrix} \! \!  \in \! \mathbb{R}^{p \times p}.
\label{eq:theWk}
\end{equation}

Finally, we define a wavelet-like synthesis transformation that is a cascade of the fundamental stages \eqref{eq:theWk} as
\begin{equation}
\mathbf{W} = \mathbf{W}_1^{(p)} \cdots \mathbf{W}_{m-1}^{(p)} \mathbf{W}_m^{(p)}.
\label{eq:theW}
\end{equation}
We call this transformation wavelet-like because $\mathbf{Wx}$ applies convolutions to parts of the signal $\mathbf{x}$ at different scales (for example, see \cite{trove.nla.gov.au/work/9684108} for a description of discrete wavelet transformations from the perspective of matrix linear algebra).

\section{The proposed dictionary learning algorithms}



Dictionary learning \cite{DumitrescuBook2018} provides heuristics that approximate solutions to the following problem: given a dataset $\mathbf{Y} \in \mathbb{R}^{n \times N}$, a sparsity level $s$ and the size of the dictionary $S$ we want to create a dictionary $\mathbf{D} \in \mathbb{R}^{n \times S}$ and the sparse representations $\mathbf{X} \in \mathbb{R}^{S \times N}$ then solve
\begin{equation}
	\underset{\mathbf{D},\ \mathbf{X}}{\text{minimize}}  \  \| \mathbf{Y} - \mathbf{DX} \|_F^2 \quad \text{subject to } \mathbf{X} \text{ is $s$--sparse}.
\end{equation}
A classic approach to this problem, which we also use, is the iterative alternating optimization algorithm: keep the dictionary fixed and update the representations and vice-versa in a loop until convergence.

In this paper, we constrain the dictionary $\mathbf{D}$ to the structures previously discussed: circulant and convolutional (including unions in both cases) and wavelet-like. Our goal is to propose numerically efficient dictionary update rules for these structures. While for the general dictionary learning problem there are several online algorithms that have been proposed \cite{Mairal:2009:ODL:1553374.1553463, 7289440, Giovanneschi2018DictionaryLF} which are computationally efficient, in this paper, we consider only batch dictionary learning and we focus on the computational complexity of the dictionary update step.

\subsection{Circulant dictionary learning}

Given a dataset $\mathbf{Y} \in \mathbb{R}^{n \times N}$ and a sparsity level $s \geq 1$ for the representations $\mathbf{X} \in \mathbb{R}^{n \times N}$, the work in \cite{CDLA} introduces an efficient way of learning a circulant dictionary $\mathbf{C} \in \mathbb{R}^{n \times n}$ by approximately solving the optimization problem:
\begin{equation}
\begin{aligned}
& \underset{\mathbf{c},\ \mathbf{X}}{\text{minimize}} & & \|\mathbf{Y}-\mathbf{CX}\|_F^2 \\
& \text{subject to} &  & \| \text{vec}(\mathbf{X}) \|_{0} \leq sN,\ \mathbf{C} = \text{circ}(\mathbf{c}),\ \| \mathbf{c} \|_2 = 1.
\end{aligned}
\label{eq:cdla}
\end{equation}
For fixed $\mathbf{X}$, to update $\mathbf{c}$ we develop the objective function to
\begin{equation}
	\| \mathbf{Y} - \mathbf{CX} \|_F^2 = \| \mathbf{FY} - \mathbf{\Sigma F X} \|_F^2 = \| \mathbf{\tilde{Y}} - \mathbf{\Sigma \tilde{X}} \|_F^2,
	\label{eq:firststeps}
\end{equation}
and in order to minimize it we set
\begin{equation}
	\sigma_1 = \frac{ \mathbf{\tilde{x}}_1^H \mathbf{\tilde{y}}_1 }{ \|\mathbf{\tilde{x}}_1\|_2^2 },\sigma_k = \frac{ \mathbf{\tilde{x}}_k^H \mathbf{\tilde{y}}_k }{ \|\mathbf{\tilde{x}}_k\|_2^2 },\ \sigma_{n-k+2} = \sigma_k^*,k=2,\dots,n,
	\label{eq:optimalsolution}
\end{equation}
where $\mathbf{\tilde{y}}_k^T$ and $\mathbf{\tilde{x}}_k^T$ are the rows of $\mathbf{\tilde{Y}} = \mathbf{FY}$ and $\mathbf{\tilde{X}} = \mathbf{FX}$.

\noindent \textbf{Remark 1.} Given $\mathbf{Y} \in \mathbb{R}^{n \times N}$ and $\mathbf{X} \in \mathbb{R}^{n \times N}$ the best circulant dictionary $\mathbf{C}$ in terms of the Frobenius norm achieves
\begin{equation}
	\underset{\mathbf{c}}{\text{minimum}} \ \|  \mathbf{Y} - \mathbf{CX} \|_F^2 = \sum_{k=1}^n  \left( \| \mathbf{y}_k \|_2^2 - \frac{ | \mathbf{\tilde{x}}_k^H \mathbf{\tilde{y}}_k |^2 }{\| \mathbf{\tilde{x}}_k \|_2^2 } \right).
\end{equation}

\noindent \textit{Proof.}
Expand the objective of \eqref{eq:cdla} using the optimal \eqref{eq:optimalsolution} as
\begin{equation}
	\begin{aligned}
	\| & \mathbf{Y} - \mathbf{CX} \|_F^2 = \| \mathbf{Y} \|_F^2 + \| \mathbf{CX} \|_F^2 - 2\text{tr}(\mathbf{CXY}^H) \\
	= & \| \mathbf{Y} \|_F^2 + \| \mathbf{F}^H \mathbf{ \Sigma F X} \|_F^2 - 2\text{tr}(\mathbf{F}^H \mathbf{\Sigma  FXY}^H) \\
	= & \| \mathbf{Y} \|_F^2 + \| \mathbf{ \Sigma \tilde{X}} \|_F^2 - 2\text{tr}(\mathbf{\Sigma  \tilde{X}} \mathbf{\tilde{Y}}^H)\\
	= & \| \mathbf{Y} \|_F^2 + \sum_{k=1}^n \frac{ | \mathbf{\tilde{x}}_k^H \mathbf{\tilde{y}}_k |^2 }{\| \mathbf{\tilde{x}}_k \|_2^2 } -2 \sum_{k=1}^n \frac{ | \mathbf{\tilde{x}}_k^H \mathbf{\tilde{y}}_k |^2 }{\| \mathbf{\tilde{x}}_k \|_2^2 }\\
	= & \| \mathbf{Y} \|_F^2 - \sum_{k=1}^n \frac{ | \mathbf{\tilde{x}}_k^H \mathbf{\tilde{y}}_k |^2 }{\| \mathbf{\tilde{x}}_k \|_2^2 }  = \sum_{k=1}^n  \left( \| \mathbf{y}_k \|_2^2 - \frac{ | \mathbf{\tilde{x}}_k^H \mathbf{\tilde{y}}_k |^2 }{\| \mathbf{\tilde{x}}_k \|_2^2 } \right).
	\end{aligned}
	\label{eq:objfunction2}
\end{equation}
In the development of \eqref{eq:objfunction2} we use the definition of the Frobenius norm, the invariance of the Frobenius norm under unitary transformations, i.e., $\| \mathbf{F} \mathbf{X} \|_F = \| \mathbf{F}^H \mathbf{X} \|_F = \| \mathbf{X} \|_F$, and the cyclic permutation of the trace, i.e., $\text{tr}(\mathbf{ABC}) = \text{tr}(\mathbf{BCA})$.

By the Cauchy-Schwarz inequality we have that $| \mathbf{\tilde{x}}_k^H \mathbf{\tilde{y}}_k |^2 \leq \| \mathbf{\tilde{x}}_k \|_2^2 \| \mathbf{\tilde{y}}_k \|_2^2$ which holds with equality if and only if the rows $\mathbf{\tilde{x}}_k$ and $\mathbf{\tilde{y}}_k$ are multiples of each other. In this case, the objective function \eqref{eq:objfunction2} develops to $\|  \mathbf{Y} - \mathbf{CX} \|_F^2 = \sum_{k=1}^n \| \mathbf{y}_k \|_2^2 - \sum_{k=1}^n \| \mathbf{\tilde{y}}_k \|_2^2$. This is the only case where the circulant dictionary can reach zero representation error.

A necessary condition that the optimal circulant dictionary $\mathbf{C}$ obeys is $\| \mathbf{\sigma} \|_2^2 = \sum_{k=1}^n \frac{ | \mathbf{\tilde{x}}_k^H \mathbf{\tilde{y}}_k |^2 }{\| \mathbf{\tilde{x}}_k \|_2^4 }  = 1$, i.e., the $\ell_2$ norm of the optimal solution of \eqref{eq:cdla} is one.$\hfill \blacksquare$

In the context of dictionary learning, to obey the unit $\ell_2$ norm constraint on $\mathbf{c}$ we should normalize the optimal solution $\mathbf{\sigma} \leftarrow \| \mathbf{\sigma} \|_2^{-1}\mathbf{\sigma}$.
This is avoided because we can always group this normalizing factor with the representations $\mathbf{X}$ instead of the circulant dictionary, i.e., $\| \mathbf{\sigma} \|_2^{-1} \mathbf{C X} = \mathbf{C} ( \| \mathbf{\sigma}| \|_2^{-1} \mathbf{X} )$. This grouping is correct because the Fourier transform preserves $\ell_2$ norms and we have that $\| \mathbf{\sigma} \|_2 = \| \mathbf{c} \|_2$, i.e., all the columns of $\mathbf{C}$ are $\ell_2$ normalized after normalizing $\mathbf{\sigma}$.

The algorithm called C--DLA, first introduced in \cite{CDLA}, has low computational complexity that is dominated by the $O(n N \log n)$ computation of $\mathbf{\tilde{Y}}$ (once) and that of $\mathbf{\tilde{X}}$ (at each iteration). The calculations in \eqref{eq:optimalsolution} take approximately $2nN$ operations: there are $\frac{n}{2}$ components in $\mathbf{\sigma}$ to be computed while $\| \mathbf{\tilde{x}}_k \|_2^2$ and $\mathbf{\tilde{x}}^H \mathbf{\tilde{y}}_k$ take $2N$ operations each.

We can limit which and how many of all the possible $n$ shifts of $\mathbf{c}$ are allowed. We achieve this by ensuring that rows of $\mathbf{X}$ corresponding to unwanted shifts are zero.

\noindent \textbf{Remark 2 (Approximating linear operators by circulant matrices).} Given $\mathbf{Y} \in \mathbb{R}^{n \times n}$, let us consider the special case of \eqref{eq:cdla} when $N = n$ and we fix $\mathbf{X} = \mathbf{I}$. Now we calculate the closest, in Frobenius norm, circulant matrix to a given linear transformation $\mathbf{Y}$. Because the Frobenius norm is elementwise the optimal solution is directly $c_k = \frac{1}{n} \sum_{(i-j) \! \! \! \mod \! n = (k-1)} y_{ i j},\ k=1,\dots,n$. Unfortunately, in general, circulant matrices do not approximate all linear transformations with high accuracy. The result is intuitive since matrices have $n^2$ degrees of freedom while circulants have only $n$. Furthermore, if we add other constraints, such as orthogonality for example, the situation is even worse: \cite{OrthogonalCirculant} shows that the set of orthonormal circulants is finite and constructs it. Therefore, researchers proposed approximating a linear transformation by a product of $O(n)$ circulant, or Toeplitz, and diagonal matrices \cite{Ye2016, Huhtanen2015}.$\hfill \blacksquare$

\subsection{Union of circulant dictionary learning}
\begin{algorithm}[t]
	\caption{ \textbf{-- UCirc--DLA--SU. } \newline \textbf{Input: } The dataset $\mathbf{Y} \in \mathbb{R}^{n \times N}$, the number of circulant atoms $L$ and the sparsity $s \leq n$. \newline \textbf{Output: } The union of $L$ circulant dictionaries $\mathbf{D} \in \mathbb{R}^{n \times nL}$ as in \eqref{eq:uniondictionary} and the sparse representations $\mathbf{X} \in \mathbb{R}^{nL \times N}$ such that $\| \mathbf{Y} - \mathbf{DX} \|_F^2$ is reduced.}
	\begin{algorithmic}
		\State \textbf{1. } Initialization: compute the singular value decomposition of the dataset $\mathbf{Y} = \mathbf{U \Sigma V}^T$, set $\mathbf{c}^{(\ell)} = \mathbf{u}_\ell$ for $\ell \leq n$,  set $\mathbf{c}^{(\ell)}$ to random $\ell_2$ normalized vectors of size $n$ for $L \geq \ell > n$ and compute the representations $\mathbf{X} = \text{OMP}(\mathbf{D}, \mathbf{Y}, s)$.
		
		\State \textbf{2. } Compute Fourier transform $\mathbf{\tilde{Y}}$, set its first row to zero.
		
		\State \textbf{3. } For $1,\dots,K:$
		\begin{itemize}
			\item Update dictionary:
			\begin{itemize}
				\item Compute all the $L$ Fourier transforms $\mathbf{\tilde{X}}^{(\ell)}$.
				\item Construct each optimal $\{ \mathbf{\sigma}^{(\ell)} \}_{\ell=1}^L$ from \eqref{eq:theoptimizationleastsquare} separately: $\{ \sigma_1^{(\ell)} \}_{\ell=1}^L = 0$ and compute $  \{\sigma_k^{(\ell)} \}_{\ell=1}^L,\ \{ \sigma_{n-k+2}^{(\ell)} \}_{\ell=1}^L = \{ (\sigma_k^{(\ell)})^* \}_{\ell=1}^L,\ k=2,\dots,\left\lceil \frac{n}{2} \right\rceil + 1,$ by \eqref{eq:optimalsolution} and then normalize the $L$ circulants $\mathbf{\sigma}^{(\ell)} \leftarrow \| \mathbf{\sigma}^{(\ell)} \|_2^{-1} \mathbf{\sigma}^{(\ell)}$.
			\end{itemize}
			\item Update sparse representations $\mathbf{X} = \text{OMP}(\mathbf{D}, \mathbf{Y}, s)$.
		\end{itemize}
	\end{algorithmic}
\end{algorithm}

Let us now consider overcomplete dictionaries (matrices that have, significantly, more columns than rows) that are unions of circulant matrices. In particular, take a dictionary which is the union of $L$ circulants:
\begin{equation}
	\mathbf{D} = \begin{bmatrix}
				\mathbf{C}^{(1)} & \mathbf{C}^{(2)} & \dots & \mathbf{C}^{(L)}
	\end{bmatrix} \in \mathbb{R}^{n \times nL},
		\label{eq:uniondictionary}
\end{equation}
where each $\mathbf{C}^{(\ell)} = \text{circ}(\mathbf{c}^{(\ell)}) =  \mathbf{F}^H \mathbf{\Sigma}^{(\ell)} \mathbf{F},\ \mathbf{\Sigma}^{(\ell)} = \text{diag}(\mathbf{\sigma}^{(\ell)}),\ \mathbf{\sigma}^{(\ell)} = \mathbf{Fc}^{(\ell)}$, is a circulant matrix. Given training data $\mathbf{Y} \in \mathbb{R}^{n \times N}$, with this structure, the dictionary learning problem has the objective:
\begin{equation}
\begin{aligned}
	\| \mathbf{Y} & - \mathbf{DX} \|_F^2 \! = \! \| \mathbf{Y} - \begin{bmatrix}
			\mathbf{C}^{(1)} & \mathbf{C}^{(2)} & \dots & \mathbf{C}^{(L)}
	\end{bmatrix} \mathbf{X} \|_F^2 \\
			= & \|\mathbf{Y} - \begin{bmatrix}
				\mathbf{F}^H \mathbf{\Sigma}^{(1)}\mathbf{F} &  \dots & \mathbf{F}^H \mathbf{\Sigma}^{(L)} \mathbf{F}
			\end{bmatrix} \mathbf{X} \|_F^2 \\
			= & \|\mathbf{Y} - \mathbf{F}^H \begin{bmatrix}
			\mathbf{\Sigma}^{(1)}\mathbf{F} &  \dots & \mathbf{\Sigma}^{(L)} \mathbf{F}
			\end{bmatrix} \mathbf{X} \|_F^2 \\
			= & \left\|  \mathbf{F} \mathbf{Y} - \sum_{\ell=1}^{L} \mathbf{\Sigma}^{(\ell)} \mathbf{F} \mathbf{X}^{(\ell)} \right\|_F^2 \! \! = \left\| \mathbf{\tilde{Y}} - \sum_{\ell=1}^{L} \mathbf{\Sigma}^{(\ell)} \mathbf{\tilde{X}}^{(\ell)} \right\|_F^2\! \! \! \!,
\end{aligned}	
\end{equation}
where the tilde matrices indicate the Fourier transforms (taken columnwise) and the representations $\mathbf{X} \in \mathbb{R}^{nL \times N}$ are separated row-wise into $L$ continuous non-overlapping blocks of size $n$ denoted $\mathbf{X}^{(\ell)} \in \mathbb{R}^{n \times N}$. A way to update all circulant components using the Fourier transforms $\mathbf{\tilde{Y}}$ and $\mathbf{\tilde{X}}$ presents itself. Denote by $\mathbf{\tilde{y}}_k^T$ the $k^\text{th}$ row of $\mathbf{\tilde{Y}}$ and by $(\mathbf{\tilde{x}}_k^{(\ell)})^T$ the $k^\text{th}$ row of $\mathbf{\tilde{X}}^{(\ell)}$. The objective function of the dictionary learning problem separates into $k=1,\dots,\left\lceil \frac{n}{2} \right\rceil + 1$ (given real valued training data $\mathbf{Y}$) distinct least squares problems like:
\begin{equation}
	\underset{\sigma_k^{(1)},\dots,\sigma_k^{(L)}}{\text{minimize}} \quad \left\| \mathbf{\tilde{y}}_k^T - \sum_{\ell=1}^{L} \sigma_k^{(\ell)} (\mathbf{\tilde{x}}_k^{(\ell)})^T \right\|_F^2.
	\label{eq:theoptimizationleastsquare}
\end{equation}
Therefore, for a fixed $k$, the diagonal entries $(k,k)$ of all $\mathbf{\Sigma}^{(\ell)}$ (which are denoted $\sigma_k^{(\ell)}$) are updated simultaneously by solving the least squares problems \eqref{eq:theoptimizationleastsquare}. Given real-valued data, mirror relations $\sigma_{n-k+2}^{(\ell)} = (\sigma_k^{(\ell)})^*,k=2,\dots,n,$ hold analogously to \eqref{eq:optimalsolution} for all $\ell =1,\dots,L$. Notice that this formulation is just a natural extension of the one dimensional least squares problems in \eqref{eq:optimalsolution}. To compute all the components of all $\mathbf{\sigma}^{(L)}$ we solve this least squares problem $\frac{n}{2}$ times -- the computational complexity is $O(n L^2 N)$.

In comparison, the union of circulants dictionary learning method presented in \cite{CDLA}, UC--DLA, updates each circulant block $\mathbf{C}^{(\ell)}$ sequentially and separately (this can be seen as a block coordinate descent approach). The new proposed learning method, called Union of Circulant Dictionary Learning Algorithm with Simultaneous Updates (UCirc--DLA--SU), is described in Algorithm 1.

\noindent \textbf{Remark 3 (Updating an unused circulant component). }Assuming that $\mathbf{X}^{(\ell)} = \mathbf{0}_{n \times N}$, i.e., the $\ell^\text{th}$ circulant matrix is never used in the representations, then we use the update $\mathbf{c}^{(\ell)} = \underset{\mathbf{z}; \ \| \mathbf{z} \|_2^2 = 1}{\arg \max} \quad \left\|  \left( \mathbf{Y} - \sum_{i=1, i\neq \ell}^L \mathbf{C}^{(i)} \mathbf{X}^{(i)} \right)  \mathbf{z} \right\|_2^2.$
This is the block update method use in UC--DLA \cite{CDLA}. Furthermore, similarly to \cite{SKSVD}, this update could be used also when atoms of block $\ell$ have a lower contribution to the reconstruction than atoms from other blocks, i.e., $\| \mathbf{X}^{(\ell)} \|_F^2 \ll \| \mathbf{X}^{(i)} \|_F^2, \forall \ i \neq \ell$.$\hfill \blacksquare$

\subsection{Union of convolutional dictionary learning}

Analogously to \eqref{eq:uniondictionary}, we define the dictionary which is a union of $L$ convolutional matrices as
\begin{equation}
\mathbf{D}_\text{conv} = \begin{bmatrix}
	\mathbf{C}_\text{conv}^{(1)} & \mathbf{C}_\text{conv}^{(2)} & \dots & \mathbf{C}_\text{conv}^{(L)}
\end{bmatrix} \in \mathbb{R}^{p \times pL},
\label{eq:uniondictionaryconv}
\end{equation}
and the objective function to minimize with respect to this union dictionary given the fixed representations $\mathbf{X}_\text{conv} \in \mathbb{R}^{pL \times N}$ (separated into $L$ continuous non-overlapping blocks denoted $\mathbf{X}_\text{conv}^{(\ell)}  = \begin{bmatrix}
\mathbf{X}^{(\ell)} \\ \mathbf{0}_{(n-1) \times N}
\end{bmatrix}  \in \mathbb{R}^{p \times N}$) is developed as
\begin{equation}
	\begin{aligned}
		\| \mathbf{Y} - &\mathbf{D}_\text{conv}\mathbf{X}_\text{conv} \|_F^2 
	\! = \! \! \left\| \text{vec}(\mathbf{\tilde{Y}}) \! - \! \! \! \sum_{\ell=1}^{L} \text{vec}(\mathbf{\Sigma}^{(\ell)}_\text{conv} \mathbf{\tilde{X}}^{(\ell)}_\text{conv}) \right\|_F^2 \\
	= & \left\| \mathbf{\tilde{y}} - \sum_{\ell=1}^{L} \mathbf{A}^{(\ell)} \mathbf{F}_{:,1:n} \mathbf{c}^{(\ell)} \right\|_F^2
	= \left\| \mathbf{\tilde{y}} - \mathbf{B} \mathbf{c} \right\|_F^2,
	\end{aligned}
	\label{eq:unionofconvolutional}
\end{equation}
where $\mathbf{B} = \begin{bmatrix} \mathbf{A}^{(1)}  \mathbf{F}_{:,1:n} & \dots & \mathbf{A}^{(L)}  \mathbf{F}_{:,1:n} \end{bmatrix}  \in \mathbb{R}^{pN \times nL}$, $\mathbf{A}^{(\ell)} = \begin{bmatrix}
\mathbf{\tilde{x}}_1^{(\ell)} \otimes \mathbf{e}_1 & \dots & \mathbf{\tilde{x}}_p^{(\ell)} \otimes \mathbf{e}_p
\end{bmatrix} \in \mathbb{C}^{pN \times p},$ with the rows of $\mathbf{\tilde{X}}_\text{conv}^{(\ell)}$, $\mathbf{c} = \begin{bmatrix}
\mathbf{c}^{(1)} & \mathbf{c}^{(2)} & \dots & \mathbf{c}^{(L)} 
\end{bmatrix}^T \in \mathbb{R}^{nL}$ and $\mathbf{F}_{:,1:n} \in \mathbb{C}^{p \times n}$ is the $p \times p$ Fourier matrix restricted to its first $n$ columns. The solution here is given by the least squares
\begin{equation}
	\mathbf{c} = ( \mathbf{B}^H  \mathbf{B}) \backslash \mathbf{B}^H \mathbf{\tilde{y}},
	\label{eq:complex}
\end{equation}
where $\mathbf{B}^H  \mathbf{B} \in \mathbb{R}^{nL \times nL}$ is a positive definite block symmetric matrix, where each $n \times n$ block is a Toeplitz matrix like $\mathbf{T}_{\ell_1 \ell_2} = \mathbf{F}_{:, 1:n}^H  (\mathbf{A}^{(\ell_1)})^H \mathbf{A}^{(\ell_2)} \mathbf{F}_{:, 1:n}$ for the $(\ell_1,\ell_2)^\text{th}$ block -- the diagonal blocks are symmetric positive definite Toeplitz. Therefore, $\mathbf{B}^H  \mathbf{B} $ is determined by $nL + (2n-1)\frac{L(L-1)}{2} $ parameters -- the first term covers the parameters of the $L$ symmetric Toeplitz diagonal blocks and the second terms covers the parameters of all $\frac{L(L-1)}{2}$ non-diagonal Toeplitz blocks. The computational burden is highest in order to calculate the diagonals $\mathbf{W}^{(\ell_1,\ell_2)} = (\mathbf{A}^{(\ell_1)})^H \mathbf{A}^{(\ell_2)}$ with entries $w_k^{(\ell_1,\ell_2)} = (\mathbf{\tilde{x}}_k^{(\ell_1)})^H \mathbf{\tilde{x}}_k^{(\ell_2)}$, i.e., the inner product of the corresponding rows from $\mathbf{\tilde{X}}_\text{conv}^{(\ell_1)}$ and $\mathbf{\tilde{X}}_\text{conv}^{(\ell_2)}$ respectively. These calculations take $O(pL^2N)$ operations. The inverse Fourier transforms of $\mathbf{W}^{(\ell_1,\ell_2)}$ to recover the entries of $\mathbf{T}_{\ell_1 \ell_2}$ take only $O(p \log_2 p)$ operations.

\begin{algorithm}[t]
	\caption{ \textbf{-- UConv--DLA--SU. } \newline \textbf{Input: } The dataset $\mathbf{Y} \in \mathbb{R}^{p \times N}$, the number of convolutional atoms $L$, the length of $\mathbf{c}$ denoted $n$, the length of the input signals $m \geq n$ (both $n$ and $m$ are chosen such that $p = n+m-1$) and the sparsity $s \leq m$. \newline \textbf{Output: } The union of $L$ convolutional dictionaries $\mathbf{D}_\text{conv} \in \mathbb{R}^{p \times pL}$ as in \eqref{eq:uniondictionaryconv} and the sparse representations $\mathbf{X}_\text{conv} \in \mathbb{R}^{pL \times N}$ such that $\| \mathbf{Y} - \mathbf{D}_\text{conv}\mathbf{X}_\text{conv} \|_F^2$ is reduced.}
	\begin{algorithmic}
		\State \textbf{1. } Initialization: set $\mathbf{c}^{(\ell)}$ for $\ell = 1,\dots,L,$ to random $\ell_2$ normalized vectors of size $p$ with non-zeros only in the first $n$ entries and compute $\mathbf{X} = \text{OMP}(\mathbf{D}_\text{conv}, \mathbf{Y}, s)$.
		
		\State \textbf{2. } Compute Fourier transform $\mathbf{\tilde{Y}}$, set its first row to zero.
		
		\State \textbf{3. } For $1,\dots,K:$
		\begin{itemize}
			\item Update dictionary:
			\begin{itemize}
				\item Compute all the $L$ Fourier transforms $\mathbf{\tilde{X}}_\text{conv}^{(\ell)}$.
				
				\item Efficiently compute \eqref{eq:complex}:
				
				\item[$\rightarrow$] Construct $\mathbf{v} = \begin{bmatrix} \mathbf{v}^{(1)} & \dots & \mathbf{v}^{(L)} \end{bmatrix}$: for $\ell=1,\dots,L$ set $z_1^{(\ell)} = 0$ and compute $z_k^{(\ell)} = (\mathbf{\tilde{x}}_k^{(\ell)})^H \mathbf{\tilde{y}}_k, \ z^{(\ell)}_{p-k+2} = (z^{(\ell)}_k)^*,\ k=2,\dots,\left\lceil \frac{p}{2} \right\rceil + 1$ and compute the inverse Fourier transform of $\mathbf{z}^{(\ell)}$ and keep only its first $n$ entries: $\mathbf{v}^{(\ell)} = \mathbf{F}_{:,1:n}^H \mathbf{z}^{(\ell)}$.
				
				\item[$\rightarrow$] Explicitly construct $\mathbf{B}^H \mathbf{B}$: for $\ell_1 = 1,\dots,L$ and $\ell_2 = \ell_1,\dots,L$ compute first column and row of the block $\mathbf{T}_{\ell_1 \ell_2}$ ($\mathbf{T}_{\ell_2 \ell_1} = \mathbf{T}_{\ell_1 \ell_2}^T$) by the inverse Fourier transform of $w_k^{(\ell_1,\ell_2)} = (\mathbf{\tilde{x}}_k^{(\ell_1)})^H \mathbf{\tilde{x}}_k^{(\ell_2)},\ w_{p-k+2}^{(\ell_1,\ell_2)} = (w_{k}^{(\ell_1,\ell_2)})^*,\ k=1,\dots,\left\lceil \frac{p}{2} \right\rceil +1$.
				
				\item[$\rightarrow$] Get $\mathbf{c}$, solve \eqref{eq:complex} by the Cholesky decomposition and normalize the $L$ convolutions $\mathbf{c}^{(\ell)} \leftarrow \| \mathbf{c}^{(\ell)} \|_2^{-1} \mathbf{c}^{(\ell)}$.
			\end{itemize}
			\item Update sparse representations $\mathbf{X} \! \! = \! \text{OMP}(\mathbf{D}_\text{conv},\! \mathbf{Y}, s)$.
		\end{itemize}
	\end{algorithmic}
\end{algorithm}
The inverse problem in \eqref{eq:complex} can be solved exactly in $O(n^3 L^3)$ via block Cholesky factorization \cite[Chapter~4.2]{Golub1996}. When $nL$ is large, an alternative approach is to use some iterative procedure like the Conjugate Gradient approach (already used in convolutional problems \cite{Wohlberg2016}) where the computational burden falls on computing matrix-vector products with the matrix $\mathbf{B}^H \mathbf{B}$ which take only $O(pL^2)$ operations. The vector $\mathbf{v} = \mathbf{B}^H \mathbf{\tilde{y}}$ has the structure $\mathbf{v} = \begin{bmatrix} \mathbf{v}^{(1)} & \dots & \mathbf{v}^{(L)} \end{bmatrix}^T$ where $\mathbf{v}^{(\ell)} = \mathbf{F}_{:,1:n}^H \mathbf{z}^{(\ell)} \in \mathbb{R}^{n}$ and $\mathbf{z}^{(\ell)} \in \mathbb{C}^p$ with entries $z_j^{(\ell)} = (\mathbf{\tilde{x}}_j^{(\ell)})^H \mathbf{\tilde{y}}_j$, i.e., the $j^\text{th}$ entry of the $\ell^\text{th}$ component is the inner product of the corresponding rows from $\mathbf{\tilde{X}}_\text{conv}^{(\ell)}$ and $\mathbf{\tilde{Y}}$ respectively. The cost of computing $\mathbf{v} \in \mathbb{R}^{nL}$ is $O(pLN)$ since the inverse Fourier transforms are only $O(p \log_2 p)$. Also, we need to compute once the Fourier transform of the dataset ($\mathbf{\tilde{Y}}$) and at each iteration all the $L$ Fourier transforms of the sparse representations $\mathbf{\tilde{X}}_\text{conv}^{(\ell)}$ which take $O(Np\log_2 p)$ and $O(LNp\log_2 p)$ overall operations respectively.

The new proposed learning method, called Union of Convolutional Dictionary Learning Algorithm with Simultaneous Updates (UConv--DLA--SU), is described in Algorithm 2.

The theoretical and algorithmic importance of dictionaries that are unions of convolutional matrices where the first columns have different (and potentially overlapping) supports has been highlighted in \cite{Elad2017} (see in particular the clear presentation of the convolutional sparse model in \cite[Figure~1]{Elad2017}): group together the first columns of each $\mathbf{C}_\text{conv}^{(\ell)}$ into the local dictionary $\mathbf{D}^{(1)}$ of size $p \times L$, do the same with the second columns into the local dictionary $\mathbf{D}^{(2)}$ and so on until $\mathbf{D}^{(p)}$. These dictionaries are called local because they are localized to the reconstruction of only a few (depending on the size of the support $n$) grouped entries from a signal, as compared to a global signal model. 

Notice that \eqref{eq:unionofconvolutional} makes use of the matrix $\mathbf{F}_{:,1:n}$ due to the sparsity pattern in $\mathbf{c}_\text{conv}$, i.e., only the first $n$ entries are non-zero. This can be generalized so that if the support of $\mathbf{c}_\text{conv}$ is denoted by $\mathcal{S}(\mathbf{c}_\text{conv})$ then the least squares problem \eqref{eq:unionofconvolutional} can be solved only on this support my making use of $\mathbf{F}_{:,\mathcal{S}(\mathbf{c}_\text{conv})} \in \mathbb{C}^{p \times |\mathcal{S}(\mathbf{c}_\text{conv})|}$, i.e., the Fourier matrix of size $p \times p$ restricted to the columns indexed in $\mathcal{S}(\mathbf{c}_\text{conv})$. Alternatively, if we do not want to (or cannot) decide the support a priori, we can add an $\ell_1$ penalty to \eqref{eq:unionofconvolutional} to promote sparse solutions.

Without the sparsity constraints in $\mathbf{c}_\text{conv}$, UConv--DLA--SU essentially reduces to Ucirc--DLA--SU but with a major numerical disadvantage: the decoupling that allows for \eqref{eq:theoptimizationleastsquare} is no longer valid and therefore the least squares problem \eqref{eq:complex} is significantly larger and harder to solve. This extra computational effort seems unavoidable if we want to impose the sparsity in $\mathbf{c}_\text{conv}$. This motivates us to discuss the possibility of updating each convolutional dictionary sequentially, just like \cite{CDLA} does for circulant dictionaries.

\noindent\textbf{Remark 4 (The special case of $L = 1$ -- the computational simplifications of constructing a single convolutional dictionary).} Following a similar development as \cite[Remark 2]{CDLA}, consider again the objective function of \eqref{eq:cdla} and develop
\begin{equation}
\|\mathbf{Y}-\mathbf{C}_\text{conv} \mathbf{X}_\text{conv} \|_F^2 
=  \| \mathbf{\tilde{y}} - \mathbf{A}\mathbf{F}_{:,1:n} \mathbf{c}  \|_F^2,
\label{eq:lessefficient}
\end{equation}
where we have defined $\mathbf{A} = \begin{bmatrix}
\mathbf{\tilde{x}}_1 \otimes \mathbf{e}_1 & \dots & \mathbf{\tilde{x}}_p \otimes \mathbf{e}_p
\end{bmatrix} \in \mathbb{C}^{pN \times p}$,
with $\{ \mathbf{e}_k \}_{k=1}^p$ the standard basis for $\mathbb{R}^p$, i.e., $\mathbf{A}$ is composed of columns from $(\mathbf{\tilde{X}}_\text{conv}^T \otimes \mathbf{I})$ corresponding to the non-zero diagonal entries of $\mathbf{\Sigma}_\text{conv}$ (see also the Khatri-Rao product \cite{KhatriRao}).

By the construction in \eqref{eq:convolution} only the first $n$ entries of $\mathbf{c}_\text{conv}$ are non-zero and therefore the optimal minimizer of \eqref{eq:lessefficient} is
\begin{equation}
\mathbf{c} =  (\mathbf{A}\mathbf{F}_{:,1:n})\backslash \mathbf{\tilde{y}}.
\label{eq:convreal}
\end{equation}

The large matrix $\mathbf{AF}_{:,1:n}$ is never explicitly constructed in the computation of $\mathbf{c} =  (\mathbf{F}_{:,1:n}^H \mathbf{A}^H \mathbf{A} \mathbf{F}_{:,1:n})^{-1} \mathbf{F}_{:,1:n}^H \mathbf{A}^H \mathbf{\tilde{y}}  =  (\mathbf{F}_{:,1:n}^H \mathbf{W} \mathbf{F}_{:,1:n})^{-1}  \mathbf{v}$, where we have defined
\begin{equation}
\mathbf{W} = \text{diag}(\begin{bmatrix}
\| \mathbf{\tilde{x}}_1 \|_2^2 & \dots & \| \mathbf{\tilde{x}}_p \|_2^2
\end{bmatrix}),\ \mathbf{v} = \mathbf{F}_{:,1:n}^H \mathbf{A}^H \mathbf{\tilde{y}}.
\label{eq:thevariables}
\end{equation}
Observe that $\mathbf{T} = \mathbf{F}_{:,1:n}^H \mathbf{W} \mathbf{F}_{:,1:n}$ is a real-valued symmetric positive definite Toeplitz matrix -- it is the upper left (the leading principal) submatrix of the $p \times p$ circulant matrix $\mathbf{F}^H \mathbf{W} \mathbf{F}$. The matrix $\mathbf{T}$ is never explicitly computed, but its first column is contained in the $n$ entries of the vector $\mathbf{F}^H \text{diag}(\mathbf{W})$. Also, notice that $\mathbf{A}^H \mathbf{\tilde{y}}$ is a vector whose $j^\text{th}$ entry is the inner product of the corresponding rows from $\mathbf{\tilde{X}}$ and $\mathbf{\tilde{Y}}$ respectively, i.e., $\mathbf{\tilde{x}}_j^H \mathbf{\tilde{y}}_j$.

The computation of $\mathbf{W}$ and $\mathbf{v}$ take approximately $O(pN)$ operations each -- because $\mathbf{X}$ is real-valued, the Fourier transforms exhibit symmetries and only half the $\ell_2$ norms in $\mathbf{W}$ and of the entries in $\mathbf{A}^H \mathbf{\tilde{y}}$ need to be computed. These calculations dominate the computational complexity since they depend on the size of the dataset $N \gg p$ -- together with the computations of the Fourier transforms $\mathbf{\tilde{X}}_\text{conv}$ (at each iteration) and $\mathbf{\tilde{Y}}$ (only once) which take $O(LN p\log_2 p)$ and $O(N p\log_2 p)$ operations respectively. The least squares problem \eqref{eq:convreal} is solved via the Levinson-Durbin algorithm \cite[Section 4.7.3]{Golub1996}, whose complexity is $4n^2$ instead of the regular $O(n^3)$ computational complexity for unstructured inverse problems.

Also, observe that this least squares solution when applied to minimizing \eqref{eq:firststeps} leads to the same optimal solution from \eqref{eq:optimalsolution}: $\mathbf{c} = (\mathbf{F}^H \mathbf{A}^H \mathbf{A} \mathbf{F})^{-1} \mathbf{F}^H  \mathbf{A}^H \mathbf{\tilde{y}} = \mathbf{F}^H \mathbf{W}^{-1} \mathbf{F} \mathbf{F}^H \mathbf{A}^H \mathbf{\tilde{y}}
= \mathbf{F}^H \mathbf{W}^{-1} \mathbf{A}^H \mathbf{\tilde{y}} = \mathbf{F}^H \mathbf{\sigma}$, with $\mathbf{W} = \text{diag}(\begin{bmatrix}
\| \mathbf{\tilde{x}}_1 \|_2^2 & \dots &  \| \mathbf{\tilde{x}}_n \|_2^2
\end{bmatrix}) $. The approach in \eqref{eq:optimalsolution} is preferred to \eqref{eq:convreal} since in the Fourier domain the overall problem is decoupled into a series of smaller size independent subproblems that can be efficiently solved in parallel.

Therefore, each single convolutional dictionary can be updated efficiently and an algorithm in the style of UC--DLA \cite{CDLA} can be proposed whenever running time is of concern or the dataset is large. In this case, because the convolutional components would not be updated simultaneously, we would expect worse results on average.$\hfill \blacksquare$

There are several places where structures related to unions of circulant and convolutional dictionaries appear. We briefly discuss next Gabor frames and then, in the following section, wavelet-like dictionaries.

\noindent \textbf{Remark 5 (A special case of time-frequency synthesis dictionary).} Union of circulant matrices also show up when studying discrete time-frequency analysis/synthesis matrices \cite{RecentGabor}. Consider the Gabor synthesis matrix given by
\begin{equation}
\mathbf{G} = \begin{bmatrix}
\mathbf{D}^{(1)} \mathbf{C} & \mathbf{D}^{(2)} \mathbf{C} & \dots & \mathbf{D}^{(m)} \mathbf{C}
\end{bmatrix} \in \mathbb{C}^{m \times m^2},
\end{equation}
where $\mathbf{C} = \text{circ}(\mathbf{g})$ for $\mathbf{g} \in \mathbb{C}^m$ which is called the Gabor window function, i.e., the matrix $\mathbf{G}$ contains circular shifts and modulations of $\mathbf{g}$. The matrices $\mathbf{D}^{(\ell)},\ell=1,\dots,m,$ are diagonal with entries $d^{(\ell)}_{kk} = \omega^{(\ell-1)(k-1)}$ and $\omega = e^{2\pi i / m}$.

In the context of compressed sensing with structured random matrices, Gabor measurement matrices have been used for sparse signal recovery \cite{TFCS}: the Gabor function of length $m$ is chosen with random entries independently and uniformly distributed on the torus $\{ z \in \mathbb{C}\ | \ |z| = 1 \}$ \cite[Theorem~2.3]{TFCS}. Here, our goal is to learn the Gabor function $\mathbf{g}$ from a given dataset $\mathbf{Y}$ such that the dictionary $\mathbf{G}$ allows good sparse representations. Now the objective function develops to:
\begin{equation}
\begin{aligned}
\|\mathbf{Y}& - \mathbf{GX} \|_F^2  = \| \mathbf{Y} -  \mathbf{D}(\mathbf{I} \otimes \mathbf{C}) \mathbf{X} \|_F^2 \\
= & \| \mathbf{y} - (((\mathbf{I} \otimes \mathbf{F})\mathbf{X})^T  \otimes \mathbf{D}(\mathbf{I} \otimes \mathbf{F}^H) ) \text{vec}(\mathbf{I} \otimes \mathbf{\Sigma}) \|_F^2 \\
= & \left\| \mathbf{y} - \begin{bmatrix}
\sum \limits_{\ell=1}^m \mathbf{\tilde{x}}^{(\ell)}_1 \otimes \mathbf{\tilde{d}}_1^{(\ell)} & \! \dots \! &  \sum \limits_{\ell=1}^m \mathbf{\tilde{x}}^{(\ell)}_m \otimes \mathbf{\tilde{d}}_m^{(\ell)}
\end{bmatrix} \mathbf{\sigma} \right\|_F^2 \\
= & \| \mathbf{y} - \mathbf{A\sigma} \|_F^2 = \| \mathbf{y} - \mathbf{AFg} \|_F^2,
\end{aligned}
\end{equation}
where we have $\mathbf{A} \in \mathbb{C}^{mN \times m}$ and we have denoted $\mathbf{y} = \text{vec}(\mathbf{Y})$, $\mathbf{D} =\begin{bmatrix}
\mathbf{D}^{(1)} & \dots & \mathbf{D}^{(m)}
\end{bmatrix} $, $(\mathbf{\tilde{x}}^{(\ell)}_k)^T$ is the $k^\text{th}$ row of $\mathbf{\tilde{X}}^{(\ell)}$ and $\mathbf{\tilde{d}}^{(\ell)}_k$ is the $k^\text{th}$ column of $\mathbf{D}^{(\ell)} \mathbf{F}^H$. Similarly to the convolutional dictionary learning case, Gabor atoms typically have compact support and we can add the sparse structure to $\mathbf{g}$ (the support of size $n \leq m$) and find the minimizer by solving the least squares problem which this time is unstructured (there is no Toeplitz structure).$\hfill \blacksquare$

\subsection{Wavelet-like dictionary learning}

Starting from \eqref{eq:theW}, in the spirit of dictionary learning, our goal is to learn a transformation from given data such that it has sparse representations. The strategy we use is to update each $\mathbf{W}_k^{(p)}$ (actually, the $\mathbf{C}_k^{(p)}$ component) while keeping all the other transformations fixed. Therefore, for the $k^\text{th}$ component we want to minimize
\begin{equation}
	\begin{aligned}
	& \| \mathbf{Y} - \mathbf{WX} \|_F^2 = \| \mathbf{Y} -  \mathbf{W}_\text{A} \mathbf{W}_k^{(p)} \mathbf{W}_\text{B} \mathbf{X} \|_F^2 \\
	= & \left\|  \mathbf{Y} -  \begin{bmatrix} \mathbf{W}_{\text{A}, 1} & \mathbf{W}_{\text{A}, 2}
				\end{bmatrix} \begin{bmatrix}
	\mathbf{C}_k^{\left(\frac{p}{2^{k-1}}\right)} & \mathbf{0} \\
	\mathbf{0} & \mathbf{I}
	\end{bmatrix} \begin{bmatrix} \mathbf{\bar{X}}_1 \\ \mathbf{\bar{X}}_2 \end{bmatrix} \right\|_F^2 \\
	= & \| \mathbf{\bar{Y}} - \mathbf{W}_{\text{A}, 1}\mathbf{C}_k^{\left(\frac{p}{2^{k-1}}\right)}\mathbf{\bar{X}}_1  \|_F^2 \\
	= & \| \mathbf{\bar{Y}} - \mathbf{W}_{\text{A}, 1} \begin{bmatrix}
			\mathbf{G}_k \mathbf{S} &  \mathbf{H}_k \mathbf{S}
	\end{bmatrix} \mathbf{\bar{X}}_1  \|_F^2 \\
	= & \| \mathbf{\bar{Y}} - \mathbf{W}_{\text{A}, 1} \mathbf{F}^H \begin{bmatrix}
				\mathbf{\Sigma}_{\mathbf{g}_k} & \mathbf{\Sigma}_{\mathbf{h}_k}
	\end{bmatrix} (\mathbf{I}_2 \otimes \mathbf{FS})
	 \mathbf{\bar{X}}_1 \|_F^2\\
	= & \| \mathbf{\bar{y}} \! - \! (((\mathbf{I}_2 \! \otimes \! \mathbf{FS}) \mathbf{\bar{X}}_1 )^T \! \! \! \otimes \! \! \mathbf{W}_{\text{A}, 1} \mathbf{F}^H) \text{vec}(\! \begin{bmatrix}
	\mathbf{\Sigma}_{\mathbf{g}_k} & \! \! \! \! \mathbf{\Sigma}_{\mathbf{h}_k}
	\end{bmatrix}\! ) \|_F^2 \\
	= & \left\| \mathbf{\bar{y}} - \mathbf{A} \begin{bmatrix}
						\mathbf{Fg}_k \\
						\mathbf{Fh}_k
	\end{bmatrix}  \right\|_F^2
	= \left\| \mathbf{\bar{y}} -  \mathbf{A} (\mathbf{I}_2 \otimes \mathbf{F}) \begin{bmatrix}
			\mathbf{g}_{k} \\
			\mathbf{h}_{k}
	\end{bmatrix}  \right\|_F^2,
	\end{aligned}
	\label{eq:waveletoptimization}
\end{equation}
where $\mathbf{F}$ is the Fourier matrix of size $\frac{p}{2^{k-1}}$, we denoted $\mathbf{W}_{\text{A}} = \mathbf{W}_1^{(p)} \cdots \mathbf{W}_{k-1}^{(p)}$, $\mathbf{W}_\text{B} = \mathbf{W}_{k+1}^{(p)} \cdots \mathbf{W}_m^{(p)}$ and $\mathbf{\bar{X}} =  \mathbf{W}_\text{B} \mathbf{X} $, $\mathbf{\bar{Y}} = \mathbf{Y} - \mathbf{W}_{\text{A}, 2} \mathbf{\bar{X}}_2$, $\mathbf{\bar{y}} = \text{vec}(\mathbf{\bar{Y}})$, $\mathbf{W}_{\text{A}, 1}$ are the first $\frac{p}{2^{k-1}}$ columns of $\mathbf{W}_{\text{A}}$, $\mathbf{\bar{X}}_1$ are the first $\frac{p}{2^{k-1}}$ rows of $\mathbf{\bar{X}}$. We have also denote $\mathbf{\Sigma}_{\mathbf{g}_k} = \text{diag}(\mathbf{Fg}_k)$ and similarly for $\mathbf{\Sigma}_{\mathbf{g}_k}$. The matrix $\mathbf{A} \in \mathbb{C}^{pN \times \frac{2p}{2^{k-1}}}$ is made up of a subset of the columns from $((\mathbf{I}_2 \otimes \mathbf{FS}) \mathbf{\bar{X}}_1 )^T \otimes \mathbf{W}_{\text{A}, 1} \mathbf{F}^H$ corresponding to the non-zero entries from $\text{vec}(\begin{bmatrix}
\mathbf{\Sigma}_{\mathbf{g}_k} & \mathbf{\Sigma}_{\mathbf{h}_k}
\end{bmatrix})$. Minimizing the quantity in \eqref{eq:waveletoptimization} leads to a least squares problem where both $\mathbf{g}_k$ and $\mathbf{h}_k$ have a fixed non-zero support of known size $n$. It obeys $n \leq \frac{p}{2^{m-1}}$ such that the circulants for $\mathbf{C}_m^{(p)}$ can be constructed.

Similarly to the union of circulants cases described before, some computational benefits arise when minimizing \eqref{eq:waveletoptimization}, i.e., computing $(\mathbf{I}_2 \otimes \mathbf{F}^H) \mathbf{A}^H \mathbf{A} (\mathbf{I}_2 \otimes \mathbf{F})\backslash (\mathbf{I}_2 \otimes \mathbf{F}^H) \mathbf{A}^H \mathbf{\bar{y}}$. For convenience we will denote $\mathbf{Q} = (((\mathbf{I}_2 \! \otimes \! \mathbf{FS}) \mathbf{\bar{X}}_1 )^T $ and $\mathbf{R} =  \mathbf{W}_{\text{A}, 1} \mathbf{F}^H$, such that $\mathbf{A} = \mathbf{Q} \odot \mathbf{R}$. Notice that $\mathbf{A}^H \mathbf{A} = \frac{p}{2^{k-1}} \begin{bmatrix}
	\mathbf{D}^{(1)} & \mathbf{D}^{(2)*} \\
	\mathbf{D}^{(2)} & \mathbf{D}^{(3)}
\end{bmatrix}$, where the blocks are diagonal with entries $d_{ii}^{(1)} = \| \mathbf{q}_i \|_2^2 \| \mathbf{r}_i \|_2^2$, $d_{ii}^{(3)} =  \| \mathbf{q}_{ \frac{p}{2^{k-1}} +i } \|_2^2 \| \mathbf{r}_i \|_2^2$ and $d_{ii}^{(2)} =  \mathbf{q}_{ \frac{p}{2^{k-1}} +i }^H \mathbf{q}_i \| \mathbf{r}_i \|_2^2$ where $\mathbf{q}_i$ and $\mathbf{r}_i$ are columns of $\mathbf{Q}$ and $\mathbf{R}$, respectively, $i = 1,\dots,\frac{p}{2^{k-1}}$. Therefore, $(\mathbf{I}_2 \otimes \mathbf{F}^H) \mathbf{A}^H \mathbf{A} (\mathbf{I}_2 \otimes \mathbf{F})$ is a $2 \times 2$ block matrix whose blocks are real-valued circulant matrices (and the diagonal blocks are also symmetric). Also, because $\mathbf{A}^H \mathbf{A}$ is symmetric positive definite it allows for a Cholesky factorization $\mathbf{L}\mathbf{L}^T$ where the matrix $\mathbf{L}$ has only the main diagonal and the secondary lower diagonal of size $\frac{p}{2^{k-1}}$ of non-zero values. A further computational benefit comes from when $n \ll p$ and we solve a least squared problem in $2n$ variables, as compared to $2p$, i.e., $\mathbf{A} (\mathbf{I}_2 \otimes \mathbf{F}) \begin{bmatrix}
	\mathbf{g}_{k} \\
	\mathbf{h}_{k}
\end{bmatrix} = \mathbf{A} (\mathbf{I}_2 \otimes \mathbf{F}_{:,1:n}) \begin{bmatrix}
\mathbf{\bar{g}}_{k} \\
\mathbf{\bar{h}}_{k}
\end{bmatrix}$, with both $\mathbf{\bar{g}}_{k}, \mathbf{\bar{h}}_{k} \in \mathbb{R}^n$. Finally, notice that $(\mathbf{I}_2 \otimes \mathbf{F}_{:,1:n}^H) \mathbf{A}^H \mathbf{A} (\mathbf{I}_2 \otimes \mathbf{F}_{:,1:n})$ has Toeplitz blocks, like in the case of UConv--DLA--SU.

The linear transformation \eqref{eq:theW} has two major advantages: i) the computational complexity of matrix-vector multiplications $\mathbf{Wx}$ with a fixed $\mathbf{x} \in \mathbb{R}^p$ is controlled by the number of stages $m$ and the length of the filters $n$, instead of the fixed matrix-vector multiplication complexity which is $O(p^2)$ and ii) it allows for learning atoms that capture features from the data at different scales, i.e., atoms of different sparsity levels.

The new proposed procedure, called Wavelet-like Dictionary Learning Algorithm (W--DLA), is described in Algorithm 3.

\noindent \textbf{Remark 6 (Extending and constraining the wavelet-like structure).} Unlike wavelets that use the same filters $\mathbf{g}$ and $\mathbf{h}$ (known as the low and high pass filters, respectively) at each stage $k$ of the transformation, we learn different filters $\mathbf{g}_k$ and $\mathbf{h}_k$. Also, the support of the filters (and their size) can be decided dynamically at every stage and the downsampling can be replaced with a general column selection matrix $\mathbf{S}$. Note that if we allow full support then the least squares problem can be solved in the Fourier domain with the complex-valued unknowns $\mathbf{\tilde{g}}_k = \mathbf{Fg}_k$ and $\mathbf{\tilde{h}}_k = \mathbf{Fh}_k$. Furthermore, each stage in \eqref{eq:theWk} applies only to the decomposition of the left-most (so-called low-frequency) components from the previous stage. In the spirit of optimal sub-band tree structuring (also known as wavelet packet decompositions) \cite{Coifman1992}, we can propose a transformation $\mathbf{W} = \mathbf{C}_1^{(p)} \cdots \mathbf{C}_{m-1}^{(p)} \mathbf{C}_m^{(p)}$ factored into $m$ stages all of the form \eqref{eq:theWn} (where again each stage has its own filters $\mathbf{g}_k$ and $\mathbf{h}_k$ which now can have support $n \leq p$) and use the same optimization procedure. Lastly, we could also propose a structured transformation similar to \eqref{eq:theWn} but based on more than two filters, e.g., $\mathbf{C}_k^{(p)} = \begin{bmatrix}
	\mathbf{G}_k \mathbf{S} & \mathbf{H}_k \mathbf{S} & \mathbf{J}_k\mathbf{S}
\end{bmatrix} \in \mathbb{R}^{p \times p}$, for a new selection (downsampling by 3) matrix $\mathbf{S} \in \mathbb{R}^{p \times \frac{p}{3}}$. Heuristics to choose $m$, $n$ (maybe at each stage $m$, i.e., having $n_k$), the location of the $n$ non-zero entries in each filter, the structure and sizes of $\mathbf{C}_k$ and $\mathbf{S}$ may be proposed to further improve the accuracy of the algorithm (or the trade-off between numerical efficiency and accuracy in terms of the representation error).

\begin{algorithm}[t]
	\caption{ \textbf{-- W--DLA. } \newline \textbf{Input: } The dataset $\mathbf{Y} \in \mathbb{R}^{p \times N}$ such that $2^m$ divides $p$ exactly, the number of stages $m \leq \log_2 p$, the size of the support of the filters denoted $n \leq \frac{p}{2^{m-1}}$ and the sparsity $s \leq p$. \newline \textbf{Output: } The wavelet-like dictionary $\mathbf{W} \in \mathbb{R}^{p \times p}$ as in \eqref{eq:theW}, the diagonal $\mathbf{D} \in \mathbb{R}^{p \times p}$ such that $\mathbf{WD}$ has unit $\ell_2$ norm columns and the sparse representations $\mathbf{X} \in \mathbb{R}^{p \times N}$ such that $\| \mathbf{Y} - \mathbf{WD}\mathbf{X} \|_F^2$ is reduced.}
	\begin{algorithmic}
		\State \textbf{1. } Initialization: set all stages $\mathbf{C}_k^{(n)} = \mathbf{I},\ k = 1,\dots,m$; compute the singular value decomposition of the dataset $\mathbf{Y} = \mathbf{U \Sigma V}^T$ and compute the sparse representations $\mathbf{X} = \mathcal{T}_s(\mathbf{U}^T \mathbf{Y})$, i.e., project and keep the $s$ largest entries in magnitude for each element (column) in the dataset $\mathbf{Y}$.
		
		\State \textbf{2. } For $1,\dots,K:$
		\begin{itemize}
			\item Update dictionary: with all other components fixed, update only the $k^\text{th}$ non-trivial component of $\mathbf{W}$ denoted $\mathbf{C}_k^{\left( \frac{p}{2^{k-1}} \right) }$ (by computing both $\mathbf{g}_k$ and $\mathbf{h}_k$ on the support of size $n$) for each $k=1,\dots,m,$ at a time by minimizing the least squares problem \eqref{eq:waveletoptimization}.
			
			\item Update $\mathbf{D}$ such that $\mathbf{WD}$ has unit $\ell_2$ norm columns and update sparse representations $\mathbf{X} = \text{OMP}(\mathbf{WD}, \mathbf{Y}, s)$.
		\end{itemize}
	\end{algorithmic}
\end{algorithm}
Orthonormal wavelets, i.e., in our case meaning that $\mathbf{W}$ and all $\mathbf{C}_k^{(p)}$ are orthonormal, are also extensively used in many application. With these orthogonality constraints the objective function develops into a simpler form than \eqref{eq:waveletoptimization} as
\begin{equation}
\begin{aligned}
\| & \mathbf{Y} - \mathbf{W}_\text{A}  \mathbf{W}_k^{(p)} \mathbf{W}_\text{B} \mathbf{X}  \|_F^2 \! = \! \| \mathbf{W}_\text{A}^T \mathbf{Y} \! - \! \mathbf{W}_k^{(p)} \mathbf{W}_\text{B} \mathbf{X} \|_F^2\\
= & \| \mathbf{W}_\text{A}^T \mathbf{Y} - \mathbf{\bar{X}}_2 - \mathbf{W}^{\left(\frac{p}{2^{k-1}}\right)} \mathbf{\bar{X}}_1 \|_F^2 \\
= & \| \mathbf{W}_A^T \mathbf{Y} - \mathbf{\bar{X}}_2 - \mathbf{F}^H \begin{bmatrix}
\mathbf{\Sigma}_{\mathbf{g}_k} & \mathbf{\Sigma}_{\mathbf{h}_k}
\end{bmatrix}  (\mathbf{I}_2 \otimes \mathbf{F}) \mathbf{\bar{X}}_1 \|_F^2 \\
= & \| \mathbf{F} ( \mathbf{W}_\text{A}^T \mathbf{Y} - \mathbf{\bar{X}}_2) -  \begin{bmatrix}
\mathbf{\Sigma}_{\mathbf{g}_k} & \mathbf{\Sigma}_{\mathbf{h}_k}
\end{bmatrix}  (\mathbf{I}_2 \otimes \mathbf{FS}) \mathbf{\bar{X}}_1 \|_F^2 \\
= & \| \mathbf{\bar{y}} - (( (\mathbf{I}_2 \otimes \mathbf{FS}) \mathbf{\bar{X}}_1)^T \otimes \mathbf{I}_n) \text{vec}( \begin{bmatrix}
\mathbf{\Sigma}_{\mathbf{g}_k} & \! \! \mathbf{\Sigma}_{\mathbf{h}_k}
\end{bmatrix} ) \|_F^2,
\end{aligned}
\end{equation}
where we have now denoted $\mathbf{\bar{y}} = \text{vec}(\mathbf{F} ( \mathbf{W}_\text{A}^T \mathbf{Y} - \mathbf{\bar{X}}_2))$ and of course the Kronecker products are never explicitly built. We minimize this quadratic objective with the additional orthogonality constraint $\mathbf{g}_k^T \mathbf{h}_k = 0$ (in order to keep $\mathbf{C}_k^{(p)}$ orthogonal). Minimizing quadratic functions under quadratic equality constraints has been studied in the past and numerical procedures are available \cite{QuadraticSquared2010}. This constraint ensures orthogonality between the columns of $\mathbf{GS}$ and $\mathbf{HS}$. To ensure orthogonality among the columns of $\mathbf{GS}$ (and $\mathbf{HS}$, respectively) we can explicitly add symmetry constraints to the filter coefficients, e.g., if $\mathbf{g}_k$ has a non-zero support of size four we have $g_{k3} = g_{k1}$ and $g_{k4} = -g_{k2}$. Alternatively, both orthogonality constraints can be added leading to a quadratic optimization problem with two quadratic constraints \cite{Guu1998}.$\hfill \blacksquare$

%

\section{Experimental results}

We now discuss numerical results that show how well the proposed methods extract shift-invariant structures from data. Since we are dealing with the dictionary learning problem our goal is to build good representations of the datasets we consider. In our proposed algorithms, in the sparse recovery phase, we use the orthogonal matching pursuit (OMP) algorithm \cite{OMP}, but any other sparse approximation method could be chosen. Also, because the sparse approximation steps are numerically expensive (at least quadratic complexity in general and applied for all $N \gg n$ data points) and repetitive operations (done at each iterative step of the alternating optimization algorithms), OMP is chosen from practical considerations, as numerically efficient implementations are available (for example \cite{AKSVD}).

\begin{figure}[t]
	\centering
	\includegraphics[trim = 18 5 30 5, clip, width=0.33\textwidth]{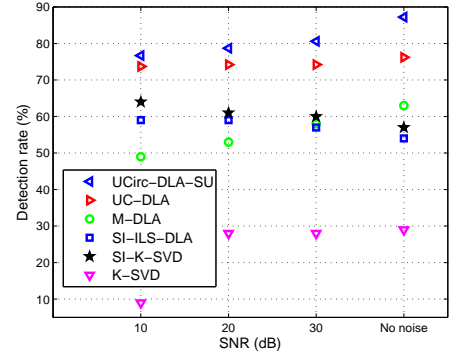}
	\caption{Average kernel recovery rate as a function of noise level for different shift invariant dictionary learning methods. We compare against SI--K--SVD \cite{AharonPhd2006}, SI--ILS--DLA \cite{SIDLA}, M--DLA \cite{Mars2012}, UC--DLA \cite{CDLA}. The K--SVD approach serves as a baseline performance indicator since it is not explicitly developed to recover shift-invariant structures.}
	\label{fig:drawSI}
\end{figure}
\begin{figure}[t]
	\centering
	\includegraphics[trim = 18 0 30 5, clip, width=0.33\textwidth]{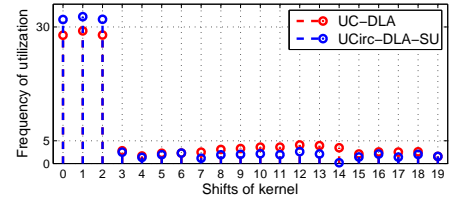}
	\caption{Average frequency of utilization for $n = 20$ atoms of the $L = 45$ circulants $\mathbf{C}_\ell$. With perfect recovery the $q = 3$ peaks should be $\frac{Ns}{Lq} \approx 44$.}
	\label{fig:drawUtilization}
\end{figure}
Notice from the description of all the proposed algorithms that each dictionary update step necessarily decrease the objective function but, unfortunately, the overall algorithms may not be monotonically convergent to a local minimum since OMP is not guaranteed in general to always reduce the objective function. As such, in this section, we also provide some experimental results where we empirically observe the converge of the proposed methods.

For datasets with a strong DC component the learned circulant dictionary might be $\mathbf{C} \approx \frac{1}{\sqrt{n}} \mathbf{1}_{n \times n}$. Therefore, preprocessing the dataset $\mathbf{Y}$ by removing the mean component is necessary and we have $\sigma_1 = 0$ in \eqref{eq:optimalsolution} since $\mathbf{\tilde{y}}_1 = \mathbf{0}_{N \times 1}$. This operation is assumed performed before the application of any algorithm developed in this paper.

\subsection{Synthetic experiments}
\begin{figure}[t]
	\centering
	\includegraphics[trim = 18 5 30 5, clip, width=0.33\textwidth]{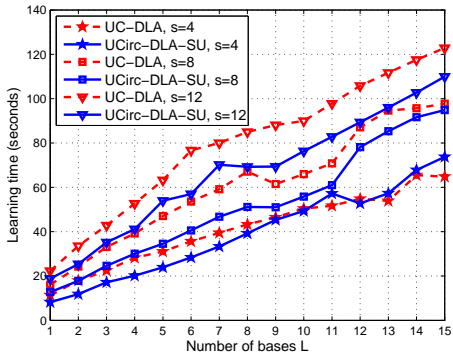}
	\caption{Average learning times for UC--DLA \cite{CDLA} and the proposed UCirc--DLA--SU over 100 random realizations. For the same parameters, the running time of UConv--DLA--SU is several times higher and therefore not shown in this plot -- this highlights the importance of solving the learning problem in the Fourier domain, when possible.}
	\label{fig:times}
\end{figure}

 We create a synthetic dataset generated from a fixed number of atoms and their shifts and the proceed to measure how well we recover them from the dictionary learning perspective. The experimental setup follows: generate $N = 2000$ signals of length $n = 20$, that are linear combinations (with sparsity $s=4$) of $L=45$ randomly generated kernel columns which are allowed to have only $q = 3$ circular shifts (out of the possible $n=20$), i.e., $\mathbf{Y} \in \mathbb{R}^{n \times N}$ where each columns is $\mathbf{y}_i = \sum_{\ell=1}^{L} \alpha_{i \ell} \mathbf{P}^{q_{i \ell}} \mathbf{c}_\ell + \mathbf{n}_i$ for $i=1,\dots,N$ with fixed $\| \mathbf{c}_\ell \|_2 = 1$, $\| \mathbf{\alpha}_i \|_0 = s$ where $\alpha_{i \ell} \in [-10, 10]$ and $q_{i \ell} \in \{ 0,\dots,q-1 \}$ are randomly uniformly distributed and $\mathbf{n}_i$ is a random Gaussian vector representing noise.
 
 First, using the synthetic dataset, we show in Figure \ref{fig:drawSI} how the UCirc--DLA--SU outperforms previously proposed methods in the task of recovering the atoms used in creating the dataset. This shows the benefit (as compared to UC--DLA) of updating all the circulant components simultaneously with each step of the algorithm. We observe that UCirc--DLA--SU achieves lower error approximately $75\%$ of the time. The typical counter-example is one where UC--DLA converges slower (in more iterations) to a slightly lower representation error, i.e., sub-optimal block calculations ultimately lead to a better final result. This observation is not surprising since both heuristic methods only approximately solve the overall original dictionary learning problem (with unknowns both $\mathbf{C}$ and $\mathbf{X}$). To show this, with the same synthetic dataset for noise level $\text{SNR} = 30\text{dB}$ in Figure \ref{fig:drawUtilization} we calculate how many times on average each atom in all circulant components (from all the $L = 45$) is used in the sparse representations. On average, UCirc--DLA--SU recovers the correct supports (in effect, the indices of the shifts used) more often than UC--DLA.
 
 Figure \ref{fig:times} shows the learning times for the union of circulants algorithms, with blocks and with simultaneous updates, for a fixed number of $K = 100$ iterations. For this test, we created a synthetic dataset $\mathbf{Y}$ of size $n = 64$ with $N = 8192$ data points and we vary the sparsity level $s$ and the number of bases in the union $L$. For $s \in \{8, 12\}$, in our experiments, UCirc--DLA--SU is always faster than UC--DLA \cite{CDLA} and the speedup is on average $20\%$ while for $s = 4$ the average speedup is only $10\%$ and for large $L$ there are cases where UCirc--DLA--SU is slower than UC--DLA. In principle, UCirc--DLA--SU should always be faster but in practice, the algorithm involves memory manipulations in order to build all the matrices for all the subproblems \eqref{eq:theoptimizationleastsquare}. This is the reason why the running time difference is not larger. Furthermore, for small $s$ the blocks used by UC--DLA are calculated rapidly (calculations are similar to the matrix operations in Remark 5) because $\mathbf{X}$ is very sparse and the overall running time is, therefore, lower.

\subsection{Experiments on ECG data}
Electrocardiography (ECG) signals \cite{ECG} have many repetitive sub-structures that could be recovered by shift-invariant dictionary learning. Therefore, in this section, we use the proposed UConv--DLA--SU to find in ECG signals short (compact support) features that are repeated. We use the MIT-BIH arrhythmia database\footnote{https://www.physionet.org/physiobank/database/mitdb/} from which we extract a normal sinus rhythm signal composed of equality length samples from five different patients, all sampled at 128 Hz. This signal is reshaped into a matrix of centered, non-overlapping sections of length $p = 64$, leading to the training dataset $\mathbf{Y} \in \mathbb{R}^{64 \times 101000}$.

\begin{figure}[t]
	\centering
	\includegraphics[trim = 15 5 25 5, clip, width=0.33\textwidth]{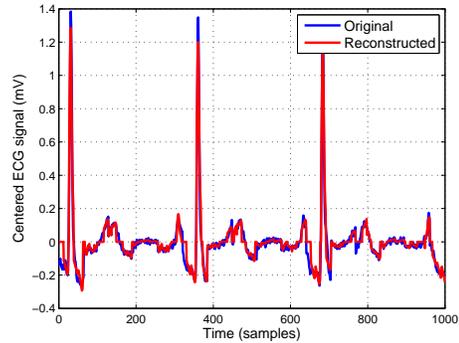}
	\caption{Original ECG sample and reconstruction by UConv--DLA--SU with support $n = 12$, sparsity $s = 4$ and $L = 2$ circulants. With these parameters, the approximation error \eqref{eq:epsilon} for the whole training dataset is $7.5\%$, showing that such data can be indeed well represented in a simple (in terms of small $L$) shift-invariant dictionary.}
	\label{fig:ecg_signals}
\end{figure}

\begin{figure}[t]
	\centering
	\includegraphics[trim = 18 0 30 5, clip, width=0.33\textwidth]{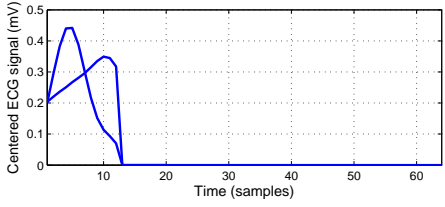}
	\caption{The $L = 2$ kernel atoms of length $p=64$ learned by UConv--DLA--SU with support $n = 12$ and sparsity $s = 4$.}
	\label{fig:ecg_atoms}
\end{figure}

Because we are searching for sub-signals with limited support, we use the UConv--DLA--SU with parameters $n = 12$, $s = 2$ and $L = 2$ to recover the shift-invariant structure. In Figure \ref{fig:ecg_signals} we show the original ECG signal and its reconstruction in the union of convolutional dictionaries. Of course, the reconstruction is not perfect but it is able to accurately capture the spikes in the data and remove some of the high-frequency features, i.e., the signal looks filtered (denoised). The second plot, Figure \ref{fig:ecg_atoms}, shows the $L=2$ learned atoms from the data which capture the spiky nature of the training signal. 

\subsection{Experiments on image data}
\begin{figure}[t]
	\centering
	\includegraphics[trim = 18 5 30 5, clip, width=0.33\textwidth]{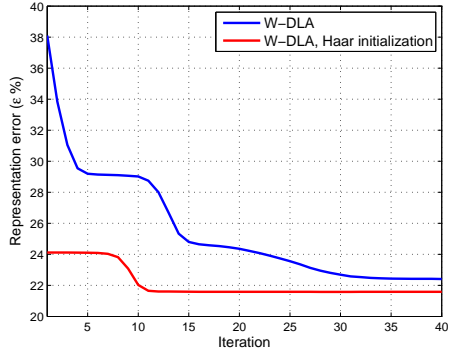}
	\caption{A typical convergence of the proposed W--DLA algorithm with the number of iterations. Since $n = 2$, $m = \log_2 p$ we also have the opportunity to initialize the filters with the Haar values.}
	\label{fig:iterations}
\end{figure}
\begin{figure}[t]
	\centering
	\includegraphics[trim = 18 5 30 45, clip, width=0.33\textwidth]{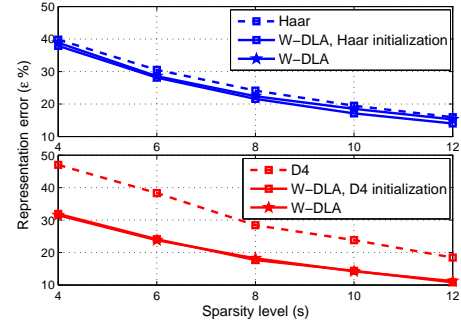}
	\caption{Representation errors as a function of the sparsity level for wavelet and W--DLA transformations. The top figure has parameters $n=2, m = \log_2 p$ and therefore allows a Haar initialization while the bottom figure has parameters $n = 4, m = -1+\log_2 p$ and allows a D4 initialization.}
	\label{fig:wavelets}
\end{figure}
\begin{figure}[t]
	\centering
	\includegraphics[trim = 18 5 30 5, clip, width=0.33\textwidth]{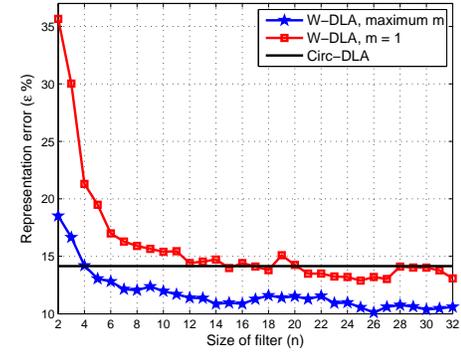}
	\caption{Representation error for W--DLA as a function of the size of the filter support $n$. For reference we show C--DLA \cite{CDLA} while W--DLA runs twice: once with fixed $m=1$ and with largest $m$ such that $n \leq \frac{p}{2^{m-1}}$ is obeyed.}
	\label{fig:wavelets_pca}
\end{figure}

The training data $\mathbf{Y}$ that we consider are taken from popular test images from the image processing literature (pirate, peppers, boat, etc.). The test dataset $\mathbf{Y} \in \mathbb{R}^{p \times N}$ consists of $8 \times 8$ non-overlapping image patches with their means removed. We consider $N = 12288$ and we have $p = 64$. To evaluate the learning algorithms, in this section we consider the relative representation error of the dataset $\mathbf{Y}$ in the dictionary $\mathbf{D}$ given the sparse representations $\mathbf{X}$ as
\begin{equation}
	\epsilon = \| \mathbf{Y} - \mathbf{DX} \|_F^2 \| \mathbf{Y} \|_F^{-2}\ (\%).
	\label{eq:epsilon}
\end{equation}

We consider image data because there are well-known wavelet transforms that efficiently encode such data. We will use the filters of the Haar and Daubechies D4 wavelet transforms, i.e., with $n=2$, $m=\log_2 p$, $\mathbf{\bar{h}}_k = \begin{bmatrix}
\frac{1}{\sqrt{2}} & -\frac{1}{\sqrt{2}}
\end{bmatrix}$, $\mathbf{\bar{g}}_k = \begin{bmatrix}
\frac{1}{\sqrt{2}}& \frac{1}{\sqrt{2}}
\end{bmatrix}$ and $n=4$, $m = -1 + \log_2 p$, $\mathbf{\bar{h}}_k = \begin{bmatrix}
\frac{1-\sqrt{3}}{4\sqrt{2}} & - \frac{3-\sqrt{3}}{4\sqrt{2}} & \frac{3+\sqrt{3}}{4\sqrt{2}} & -\frac{1+\sqrt{3}}{4\sqrt{2}} \end{bmatrix}$, $\mathbf{\bar{g}}_k = \begin{bmatrix}
\frac{1+\sqrt{3}}{4\sqrt{2}} & \frac{3+\sqrt{3}}{4\sqrt{2}} & \frac{3-\sqrt{3}}{4\sqrt{2}} & \frac{1-\sqrt{3}}{4\sqrt{2}}
\end{bmatrix}$, respectively, for all $k$. The filters are chosen such that the resulting $\mathbf{W}$ is orthonormal.

First, we show in Figure \ref{fig:iterations} the experimental convergence of the proposed W--DLA. If we also impose an orthogonality constraint on $\mathbf{W}$ then we can avoid OMP for the sparse representations and just a projection operation guarantees optimal sparse representations. The figure shows that, when available, it is convenient to initialize the W--DLA with well-known wavelet filters since the algorithm converges faster and to slightly lower representation errors. Still, the differences are not significant and wavelets do not exist for every $n,m$.

Then, in Figure \ref{fig:wavelets} we show how the representation error varies with the sparsity level $s$. We also run W--DLA initialized with the well-known wavelet filter coefficients Haar and D4, respectively. W--DLA is always able to improve the representation performance, even when starting with the wavelet coefficients. In the Haar case the improvement is small due to the small number of free filter parameters to learn, i.e., only 24: $m = \log_2 p = 6$ stages each with 2 filters and each with 2 coefficients. In the D4 case, the representation error is improved significantly. Both figures show that, when available, wavelet coefficients provide an excellent initialization even slightly better than the proposed W--DLA (also confirmed in Figure \ref{fig:iterations}). Since these wavelets are not available for all choices $n,m$ the purpose of this plot is to show that the proposed initialization provides very good results in general.

Finally, in Figure \ref{fig:wavelets_pca} we show the effect that parameters $n$ and $m$ have on the representation error. For reference, we show the representation error of C--DLA which has $p=64$ free parameters to learn. The performance of this dictionary is approximately matched by W--DLA with $n = 4$ and $m = -1 + \log_2 p$ which has $40$ free parameters to learn: $m = 5$ stages each with 2 filters of support 4 each. Notice that the representation error plateaus after $n = 8$. In general, dictionaries built with W--DLA have $2nm$ degrees of freedom. We also show a version of W--DLA where we keep $m=1$ and vary only $m$ in which case, of course, the representation error decreases. Note that in this case each run of W--DLA is initialized with a random set of coefficients. To show monotonic convergence it would help to initialize the filters of size $n$ with those previously computed of support $n-1$.

\section{Conclusions}

In this paper, we propose several algorithms that learn, under different constraints, shift-invariant structures from data. Our work is based on using circulant matrices and finding numerically efficient closed-form solutions to the dictionary update steps, by least-squares. We analyze the behavior of the algorithms on various data sources, we compare and show we outperform previously proposed algorithms from the literature.

\bibliographystyle{IEEEtran}
\bibliography{refs}

\end{document}